\documentclass[11pt]{article}

\usepackage[final]{acl}

\usepackage{times}
\usepackage{latexsym}
\usepackage{amsmath}
\usepackage{amssymb}
\usepackage{booktabs}
\usepackage{tabularx}
\usepackage[table]{xcolor}
\usepackage{caption}
\usepackage{makecell}
\definecolor{RowGray}{gray}{0.8}
\usepackage{pifont}
\usepackage{multirow}
\usepackage{algorithm}
\usepackage{algpseudocode}
\usepackage{caption}    
\usepackage{cuted}  

\usepackage[most]{tcolorbox}
\usepackage{xcolor}
\usepackage{soul} 
\sethlcolor{yellow!25}

\usepackage{microtype}
\usepackage[T1]{fontenc}

\usepackage[utf8]{inputenc}

\usepackage{microtype}

\usepackage{inconsolata}

\usepackage{graphicx}

\newtcolorbox{StoryCard}[1][]{
  enhanced,
  colback=white,
  colframe=black!65,
  boxrule=0.9pt,
  arc=3.5mm,
  left=6mm,right=6mm,top=5mm,bottom=5mm,
  fonttitle=\bfseries,
  title=#1,
  drop shadow={black!15!white},
}

\title{\textsc{\textsc{BookAgent}}: Orchestrating Safety-Aware Visual Narratives via\\Multi-Agent Cognitive Calibration}

\author{
Bo Gao \\
Carnegie Mellon University \\
\texttt{bogao@andrew.cmu.edu}
\And
Chang Liu \\
University of Science and \\
Technology of China \\
\texttt{lc980413@mail.ustc.edu.cn}
\And
Yuyang Miao \\
Imperial College London \\
\texttt{ym520@ic.ac.uk}
\AND
Siyuan Ma \\
Nanyang Technological University \\
\texttt{MASI0004@e.ntu.edu.sg}
\And
Ser-Nam Lim\thanks{Corresponding author.} \\
University of Central Florida \\
\texttt{sernam@gmail.com}
}

\begin{document}

\maketitle

\begin{abstract}
Recent advancements in Large Generative Models (LGMs) have revolutionized multi-modal generation.
However, generating illustrated storybooks remains an open challenge, where prior works mainly decompose this task into separate stages, and thus, holistic multi-modal grounding remains limited.
%
Besides, while safety alignment is studied for text- or image-only generation, existing works rarely integrate child-specific safety constraints into narrative planning and sequence-level multi-modal verification.
To address these limitations, we propose \textsc{\textsc{BookAgent}}, a safety-aware multi-agent collaboration framework designed for high-quality, safety-aware visual narratives. 
Different from prior story visualization models that assume a fixed storyline sequence, \textsc{\textsc{BookAgent}} targets end-to-end storybook synthesis from a user draft by jointly planning, scripting, illustrating, and globally repairing inconsistencies.
To ensure precise multi-modal grounding, \textsc{\textsc{BookAgent}} dynamically calibrates page-level alignment between textual scripts and visual layouts.
Furthermore, \textsc{\textsc{BookAgent}} calibrates holistic consistency from the temporal dimension, by verifying-then-rectifying global inconsistencies in character identity and storytelling logic.
Extensive experiments demonstrate that \textsc{\textsc{BookAgent}} significantly outperforms current methods in narrative coherence, visual consistency, and safety compliance, offering a robust paradigm for reliable agents in complex multi-modal creation. The implementation will be publicly released at \url{https://github.com/bogao-code/BookAgent/tree/main}.

\end{abstract}

\begin{figure}[t]
    \centering
    \includegraphics[width=0.48\textwidth]{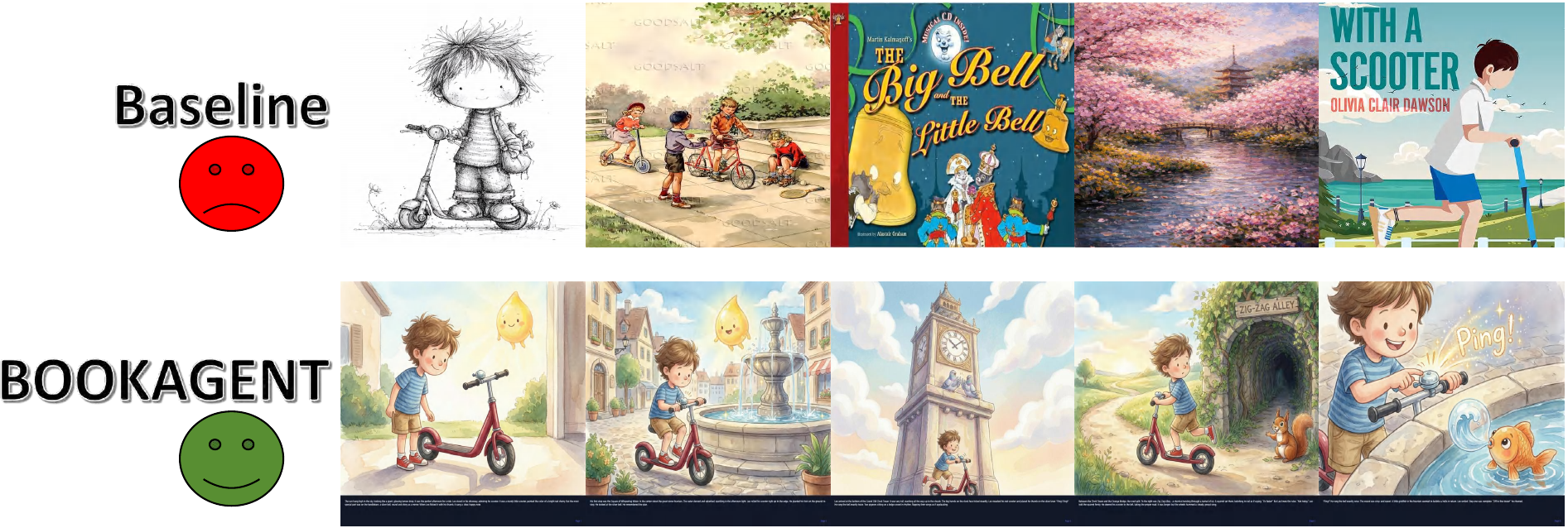}
    \caption{
    \textbf{Teaser: Long-horizon story consistency requires collaboration.}
    Given the same multi-step story prompt with strict ordering and counting constraints,
    a single-pass baseline generation fails to preserve character identity and temporal consistency across pages (top).
    In contrast, \textsc{BookAgent} leverages multi-agent collaboration to maintain stable characters, correct event order,
    and consistent visual attributes throughout the entire story sequence (bottom).
    }
    \label{fig:teaser}
    \vspace{-2em}
\end{figure}
\section{Introduction}

\begin{figure*}[t]
    \centering
    \includegraphics[width=\textwidth]{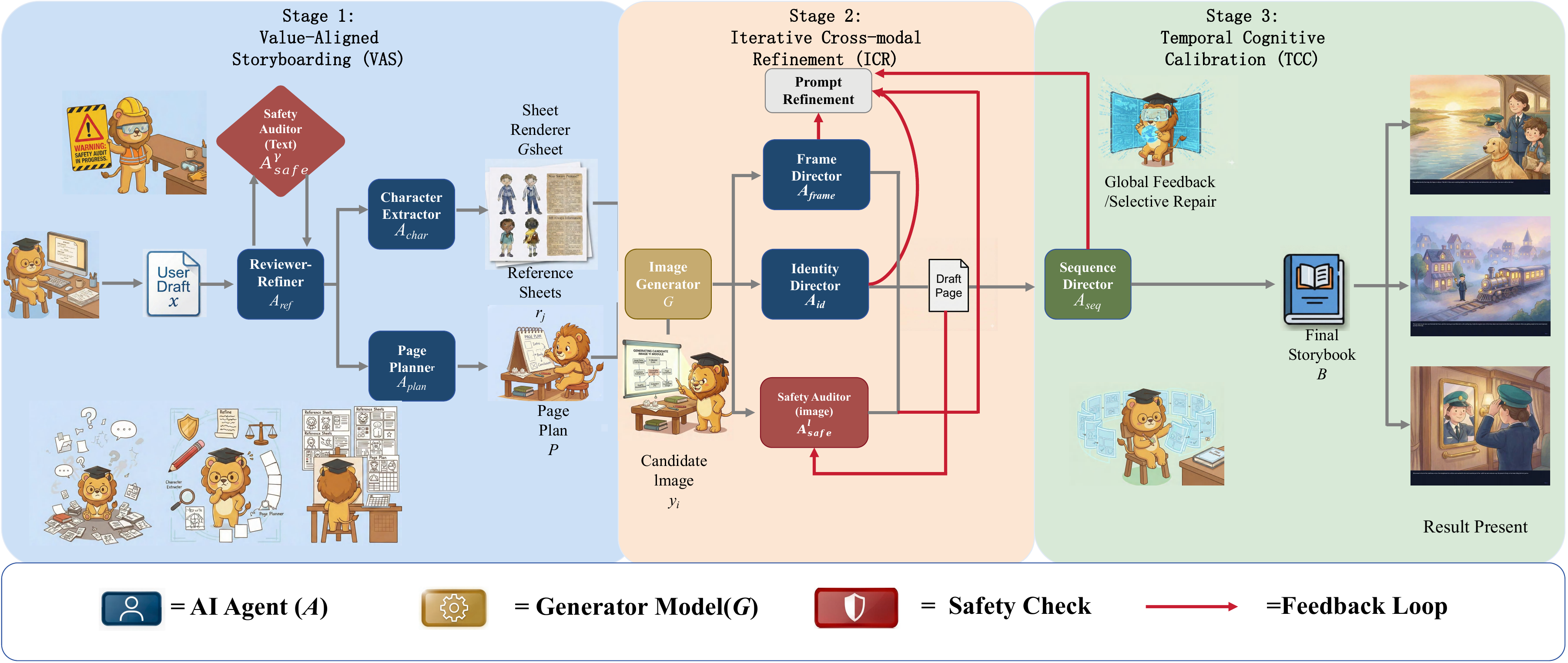}
    \caption{
    \textbf{Overview of \textsc{BookAgent}.}
    The framework follows a closed-loop, multi-agent architecture with three mechanisms.
    \textbf{Stage 1: Value-Aligned Storyboarding (VAS)} audits the input story against safety guardrails and structures it into a page plan with extracted characters and a reusable character sheet.
    \textbf{Stage 2: Iterative Cross-modal Refinement (ICR)} iteratively refines page prompts and generates candidate images, guided by frame-, identity-, and sequence-level directors with multimodal safety auditing, to improve page-level grounding and visual quality.
    \textbf{{Stage 3:Temporal Cognitive Calibration (TCC)} performs global review over the full sequence to detect and correct long-horizon inconsistencies in character identity and narrative logic.}
    }
    \label{fig:pipeline}
    \vspace{-1.5em}
\end{figure*}

Visual narratives, ranging from illustrated storybooks to complex comics, represent a fundamental medium of human communication that combines both linguistic storytelling and visual imagination.
In the era of Large Generative Models (LGMs) \cite{ddpm, sd, ddim, classifier-guidance, llama, llama2, qwen, qwen-vl}, we have witnessed astonishing capabilities in both textual and visual content generation.
The convergence of these capabilities enables the potential of translating abstract ideas into coherent, multi-modal storybooks.
However,  automating this process with existing methods is not a trivial task.
It requires an integrated system to generate coherent narrative flow, ensure semantic alignment between text and pixels, and obey strict safety standards.

LLM-based agents have recently demonstrated strong capability in decomposing complex goals into executable plans and orchestrating multi-step generation workflows in purely textual settings, e.g., by interleaving reasoning traces with tool actions \cite{react} or by learning when and how to call external APIs \cite{toolformer}.
Visual narrative tasks like storybook generation pushes such agentic reasoning into a genuinely multi-modal regime, which is normally conducted in a stage-by-stage manner by splitting the generation processes of visual and textual contents.
This process requires three key aspects to address, i.e., \textit{cross-modal alignment}, \textit{global consistency}, and \textit{safety}.
Regarding \emph{cross-modal alignment}, most existing works \cite{storydalle, intelligent-grimm} assume a given storyline sequence and produce visual content as a separate step, with more recent efforts incorporating stronger language understanding capability of LLMs into the agentic system \cite{storygpt-v}.
Despite these advances, the coupling between linguistic and visual narratives is still weak, where visual contents rarely provide structured feedback to revise the script, making bi-directional grounding and page-level calibration under-specified.
Considering \emph{global consistency}, this aspect still remains challenging beyond local alignment, as story-level generation requires long-range reasoning over entity identity, coreference, and causal relations across pages.
Some works \cite{storyimager} mainly rely on history conditioning, which can still suffer from appearance drift and role entanglement as the sequence length grows.
Therefore, explicit sequence-level verification-and-repair that jointly reasons over text, images, and multi-character coreference is expected.
\textbf{Third}, domain-specific \emph{safety} is under-explored, particularly for child-oriented storybooks.
While the safety and NLP community have put great emphasis on addressing general-purpose NSFW generation \cite{safe-clip, safegen} and child-safe text generation, respectively \cite{kidlm}
existing methods seldom integrate child-specific safety constraints into narrative planning and global consistency checking, leaving safety to generic post-hoc filters.
Ideally, a solution should function as a cohesive cognitive system, unifying the planning capability of LLM-based agents with multi-modal generators, and closing the loop with page-level verification and sequence-level refinement under explicit child-safety guardrails.

To bridge these gaps, we introduce \textsc{\textsc{BookAgent}}, a comprehensive multi-agent framework that treats storybook generation as a collaborative, safety-aware cognitive process.
Unlike previous works based on a separate paradigm that first fixes a storyline and then autoregressively produces content sequences, our approach is implemented in an end-to-end paradigm, meaning that it unifies text and image generation through a closed-loop architecture, along with three distinct mechanisms, namely Value-Aligned Storyboarding (VAS), Iterative Cross-modal Refinement (ICR), and Temporal Cognitive Calibration (TCC).
To ensure safety and value alignment of the inputs, VAS serves as the component that assists agents to rigorously audit and structure the narrative against safety guardrails before visualization begins.
ICR is the dynamic feedback loop where the system generates, evaluates, and re-generates page-level content, ensuring precise grounding between the script and the visual layout.
To enforce long-term logic, TCC performs global reasoning that reviews the entire generated sequence to identify and rectify inconsistencies in character identity and storytelling flow.
Extensive experiments indicate that \textsc{\textsc{BookAgent}} not only generates aesthetically pleasing storybooks, but also sets a new standard for narrative coherence and safety compliance.
It is worth noting that \textsc{BookAgent} is the first attempt to perform storybook content generation in an end-to-end manner, rather than in a stage-by-stage way, meaning that simultaneous multimodal content generation shows a solid reference to facilitate the inter- and intra- consistency across both modalities.

\section{Related Work}
\paragraph{Agent-based Storybook Synthesis.}
Research in cross-modal storybook synthesis has evolved from text-only planning and independent image sequence generation to recent agentic workflows \cite{reflexion, generative-agents, gorilla, critic, voyager, swe-agent, vipergpt, reasoning, tot, language-agent-tree, palm-e} that bridge reasoning with controllable synthesis.
Early efforts focused on enforcing textual coherence through hierarchical structures, where the key challenges lie at different techniques, e.g., recurrent networks \cite{storygan}, Transformer \cite{storydalle}, memory module \cite{make-a-story}, masking mechanism \cite{storyimager}, to establish the mappings between the storyline and image sequences.
With the development of agentic system \cite{react, toolformer}, all aforementioned capabilities can be effectively orchestrated into one united system.
In doing so, TaleCrafter \cite{talecrafter} combines story-to-prompt and layout generation agents; \textit{StoryGPT-V} \cite{storygpt-v} utilizes an LLM to align character descriptions with diffusion models.
Unlike these works, which often assume a fixed storyline or a one-way generation pipeline, \textsc{\textsc{BookAgent}} targets end-to-end synthesis to simultaneously produce visual and textual contents.

\paragraph{Safety-Aware Content Generation.}
Safety alignment is fundamentally vital for child-central content generation.
This particular domain-specific requirement has motivated a line of work in verifying non-toxic generation.
Specifically, Safe Latent Diffusion \cite{safe-latent-diffusion} uses language-defined safety concepts to guide sampling away from inappropriate degeneration. 
Safe-CLIP \cite{safe-clip} unlearns toxic associations in the embedding space.
DUO \cite{duo} applies preference optimization to directly unlearn unsafe features.
RECE \cite{rece} utilizes closed-form concept erasure to prevent the regeneration of erased concepts.
Different from these approaches that typically operate as post-hoc filters or single-turn constraints, \textsc{\textsc{BookAgent}} integrates safety directly into the narrative planning and sequence verification stages, enabling the early prevention of unsafe plot trajectories and the global repair of value-misaligned content in long-form narratives.

\begin{table*}[t]
\centering
\footnotesize
\caption{Role-based decomposition with fixed I/O contracts enabling verification, identity anchoring, selective repair, and child-safety enforcement.}
\renewcommand{\arraystretch}{1.12}
\setlength{\tabcolsep}{4pt}
\rowcolors{2}{RowGray}{white}

\resizebox{\textwidth}{!}{%
\begin{tabular}{@{}l l m{5.2cm} m{7.2cm}@{}}
\toprule
\textbf{Role} & \textbf{Symbol} & \textbf{I/O contract} & \textbf{Primary responsibility} \\
\midrule
Reviewer--Refiner & $\mathcal{A}_{\mathrm{ref}}$ &
\textbf{In:} $x, K, s$ \newline
\textbf{Out:} $\hat{x}$, mode, feedback &
Review the draft and either lightly polish or strongly rewrite it to match $K$ pages; improve coherence and reduce ambiguity; enforce $\le 5$ recurring characters. \\

Page Planner & $\mathcal{A}_{\mathrm{plan}}$ &
\textbf{In:} $\hat{x}, I_0, K, s$ \newline
\textbf{Out:} $\mathcal{P}=\{(t_i,p_i^{(0)})\}_{i=1}^{K}$ &
Decompose the refined story into page texts and initial prompts encoding local semantics + global style. \\

Character Extractor & $\mathcal{A}_{\mathrm{char}}$ &
\textbf{In:} $\hat{x}, s$ \newline
\textbf{Out:} $\mathcal{C}=\{c_j\}_{j=1}^{C}$ &
Extract up to $C\le 5$ recurring characters with stable ids and concise visual descriptors (species/colors/clothing) for identity anchoring. \\

Character Sheet Renderer & $\mathcal{G}_{\mathrm{sheet}}$ &
\textbf{In:} $d_j$ (visual descriptor), $s$ \newline
\textbf{Out:} $r_j$ &
Render a clean neutral-background reference sheet for each recurring character; optionally reuse user-provided inspiration image as the main character sheet. \\

Image Generator (ref-conditioned) & $\mathcal{G}$ &
\textbf{In:} $p_i^{(r)}$, refs $\mathcal{R}_i$ \newline
\textbf{Out:} $y_i^{(r)}$ &
Generate illustration candidates conditioned on the current prompt and a set of visual references (character sheets + short-term context). \\

Frame Director & $\mathcal{A}_{\mathrm{frame}}$ &
\textbf{In:} $(t_i, y_i^{(r)})$ \newline
\textbf{Out:} $\alpha_i^{(r)}$, $\Delta_i^{(r)}$ &
Verify page-level text--image faithfulness; attribute actionable issues for prompt revision. \\

Identity Director & $\mathcal{A}_{\mathrm{id}}$ &
\textbf{In:} $(t_i, y_i^{(r)}, \{r_j\}_{j=1}^{C}, s)$ \newline
\textbf{Out:} $\eta_i^{(r)}$, $\Omega_i^{(r)}$ &
Verify character identity and key recurring attributes against the reference sheets (e.g., species/color/clothing drift, missing/extra main characters). \\

Sequence Director & $\mathcal{A}_{\mathrm{seq}}$ &
\textbf{In:} $\mathcal{B}^{(m)}=\{(t_i,y_i)\}_{i=1}^{K}, s$ \newline
\textbf{Out:} $\beta^{(m)}$, $\Gamma^{(m)}$, $\mathcal{I}^{(m)}$ &
Verify cross-page continuity (identity/props/style) and attribute failures to a sparse set of problem pages for selective repair. \\

Safety Auditor (Text) & $\mathcal{A}_{\mathrm{safe}}^{T}$ &
\textbf{In:} $z\in\{x,\hat{x}\}$ \newline
\textbf{Out:} $\mathcal{S}_T(z)$, $\rho_T$ &
Audit child-safety of text; if unsafe, sanitize via constrained rewriting. \\

Safety Auditor (Image) & $\mathcal{A}_{\mathrm{safe}}^{I}$ &
\textbf{In:} $y_i^{(r)}$ \newline
\textbf{Out:} $\mathcal{S}_I(y)$, $\rho_I$ &
Audit child-safety of images; reject unsafe candidates and harden prompts with explicit safety constraints. \\
\bottomrule
\end{tabular}%
}

\label{tab:roles}
\vspace{-2.em}
\end{table*}

\section{Methodology} \label{sec:method}

\subsection{Preliminaries}
We formulate the storybook synthesis problem as a constrained optimization task.
Let $x$ denote the user-provided draft, $I_0$ an optional inspiration image, $K$ the target page count, and $s$ a global style descriptor.
Our goal is to generate a storybook $\mathcal{B} \triangleq \{(t_i, y_i)\}_{i=1}^{K}$, where $t_i$ represents the narrative script and $y_i$ the illustration for the $i$-th page.

The system is orchestrated by a set of specialized agents driven by Multimodal LLMs: a \textit{Reviewer--Refiner} $\mathcal{A}_{\mathrm{ref}}$, a \textit{Page Planner} $\mathcal{A}_{\mathrm{plan}}$, a \textit{Character Extractor} $\mathcal{A}_{\mathrm{char}}$, a \textit{Frame Director} $\mathcal{A}_{\mathrm{frame}}$, an \textit{Identity Director} $\mathcal{A}_{\mathrm{id}}$, and a \textit{Sequence Director} $\mathcal{A}_{\mathrm{seq}}$.
Visual synthesis is performed by a reference-conditioned \textit{Image Generator} $\mathcal{G}$ and a \textit{Character Sheet Renderer} $\mathcal{G}_{\mathrm{sheet}}$.
Safety is enforced by text and image auditors, formulated as:
\begin{equation}
\begin{aligned}
\mathcal{A}_{\mathrm{safe}}^{T}(\cdot) &\to (\mathcal{S}_T, \rho_T), \\
\mathcal{A}_{\mathrm{safe}}^{I}(\cdot) &\to (\mathcal{S}_I, \rho_I),
\end{aligned}
\end{equation}
where $\mathcal{S} \in \{0, 1\}$ denotes the binary safety decision and $\rho$ represents the reasoning (e.g., "violent content detected").

We aim to maximize the overall quality considering faithfulness, identity consistency, and sequence coherence, subject to hard safety constraints:
\begin{equation}
\small
\begin{aligned}
\max_{\hat{x},\,\{y_i\}_{i=1}^{K}} & \sum_{i=1}^{K} \left[ \alpha(t_i,y_i) + \eta(y_i, \{r_j\}) \right] + \lambda \beta(\mathcal{B}) \\
\text{s.t.} & \mathcal{S}_T(\hat{x})=1, \mathcal{S}_I(y_i)=1, \forall i=1,\dots,K,
\end{aligned}
\end{equation}
where $\alpha(\cdot)$ measures text--image faithfulness, $\eta(\cdot)$ measures identity consistency, and $\beta(\cdot)$ measures global sequence continuity.
We approximate this objective via a three-stage hierarchical workflow: Value-Aligned Storyboarding (VAS), Iterative Cross-modal Refinement (ICR), and Temporal Cognitive Calibration (TCC).

\subsection{Value-Aligned Storyboarding}
This stage transforms the raw draft into a structured blueprint and establishes visual anchors.
Since user drafts are often coarse, the \textit{Reviewer--Refiner} $\mathcal{A}_{\mathrm{ref}}$ rewrites the draft $x$ to match $K$ pages:
\begin{equation}
\hat{x} = \mathcal{A}_{\mathrm{ref}}(x, K, s).
\end{equation}
The output $\hat{x}$ is verified by $\mathcal{A}_{\mathrm{safe}}^{T}$; if $\mathcal{S}_T(\hat{x})=0$, the refiner utilizes the safety critique $\rho_T$ to guide constrained rewriting until standards are met.

Next, we extract recurring characters and generate canonical reference sheets prior to page generation.
The \textit{Character Extractor} $\mathcal{A}_{\mathrm{char}}$ identifies up to $C$ main characters from the refined story:
\begin{equation}
\mathcal{C} = \{c_j\}_{j=1}^{C} = \mathcal{A}_{\mathrm{char}}(\hat{x}, s),
\end{equation}
where each $c_j$ contains a stable identity and a concise visual descriptor $d_j$.
The \textit{Character Sheet Renderer} $\mathcal{G}_{\mathrm{sheet}}$ then produces neutral-background reference images:
\begin{equation}
r_j = \mathcal{G}_{\mathrm{sheet}}(d_j, s), \quad \forall j \in \{1, \dots, C\}.
\end{equation}
These reference sheets $\{r_j\}_{j=1}^{C}$ serve as the ground truth for identity verification in subsequent stages.
Finally, the \textit{Page Planner} $\mathcal{A}_{\mathrm{plan}}$ decomposes the story into a page-wise plan $\mathcal{P}$:
\begin{equation}
\mathcal{P} \triangleq \{(t_i, p_i^{(0)})\}_{i=1}^{K} = \mathcal{A}_{\mathrm{plan}}(\hat{x}, I_0, K, s),
\end{equation}
where $p_i^{(0)}$ is the initial prompt for page $i$, encoding both local semantics and global style requirements.

\subsection{Iterative Cross-modal Refinement} \label{sec:icr}
Generating high-quality storybook content requires iterative optimization. We employ a budgeted generate--verify--revise loop.
For each page $i$, we first retrieve relevant character sheets $\mathcal{R}_i = \{r_j \mid c_j \in \text{Entities}(t_i)\}$ based on the narrative.
At attempt $r < R$, we generate an image using the \textit{Image Generator} $\mathcal{G}$, formulated by:
\begin{equation}
y_i^{(r)} \sim \mathcal{G}(p_i^{(r)}, \mathcal{R}_i).
\end{equation}
We then execute a dual-branch verification. The \textit{Frame Director} $\mathcal{A}_{\mathrm{frame}}$ evaluates faithfulness, outputting score $\alpha_i^{(r)}$ and semantic issues $\Delta_i^{(r)}$. Simultaneously, the \textit{Identity Director} $\mathcal{A}_{\mathrm{id}}$ checks consistency against $\mathcal{R}_i$, yielding identity score $\eta_i^{(r)}$ and issues $\Omega_i^{(r)}$.
Afterwards, we unify these feedbacks to update the prompt $p_i^{(r+1)}$, utilizing a local memory $\mathcal{M}_i$ to accumulate historical constraints and prevent regression:
\begin{equation}
\small
p_i^{(r+1)} \!=\!
\begin{cases}
p_i^{(r)} \oplus \Psi(\rho_I), & \text{if } \mathcal{S}_I(y_i^{(r)}) = 0, \\
p_i^{(r)} \oplus \Phi(\Delta_i^{(r)}, \Omega_i^{(r)}, \mathcal{M}_i), & \text{otherwise},
\end{cases}
\end{equation}
where $\Psi(\cdot)$ converts safety reasoning $\rho_I$ into explicit negative constraints, and $\Phi(\cdot)$ aggregates current semantic/identity critiques with historical issues stored in $\mathcal{M}_i$.
We accept a candidate if it is safe, faithful ($\alpha_i^{(r)} \ge \tau_{\alpha}$), and identity-consistent ($\eta_i^{(r)} \ge \tau_{\eta}$). If the budget is exhausted, we select the best safe candidate:
\begin{equation}
\begin{aligned}
y_i = \arg\max_{y \in \{y_i^{(r)}\}}& (\alpha(t_i, y) + \eta(y, \mathcal{R}_i)) \\
\text{s.t.}& \quad  \mathcal{S}_I(y)=1.
\end{aligned}
\end{equation}

\subsection{Temporal Cognitive Calibration}
The final stage ensures cross-page consistency throughout the generated storybook.
Specifically, given the sequence $\mathcal{B}^{(m)}$ from the ICR stage, the \textit{Sequence Director} $\mathcal{A}_{\mathrm{seq}}$ performs a global audit:
\begin{equation}
(\beta^{(m)}, \Gamma^{(m)}, \mathcal{I}^{(m)}) = \mathcal{A}_{\mathrm{seq}}(\mathcal{B}^{(m)}, s),
\end{equation}
where $\Gamma^{(m)}$ contains global critiques and $\mathcal{I}^{(m)}$ is the set of indices for inconsistent pages.
If the consistency score $\beta^{(m)}$ falls below the sequence threshold $\tau_{\beta}$, we trigger a selective repair mechanism.
For each problem page $k \in \mathcal{I}^{(m)}$, we update its prompt with global context constraints derived from $\Gamma^{(m)}$ and re-enter the ICR loop (Sec. \ref{sec:icr}) with stricter reference conditioning, producing a refined book $\mathcal{B}^{(m+1)}$.
This cycle repeats until convergence or a maximum round limit is reached.

\section{Experiment}

\subsection{Experimental Setup}
\noindent\textbf{Datasets.}
Beyond standard qualitative benchmarks, we curate a specialized suite of stories designed to rigorously stress-test long-horizon visual consistency. 
Spanning from $5$ to $20$ pages, these narratives impose complex constraints that necessitate robust memory and joint reasoning. 
Specifically, the evaluation protocol enforces consistency across four rigorous dimensions. 
We first establish \textit{Identity Anchors} which bind characters to unique and non-interchangeable accessories. 
This is coupled with \textit{Symbolic Logic}, requiring exact object counts and fixed associations between color and shape. 
Additionally, \textit{Spatial Relations} mandate consistent relative positions, such as left versus right, alongside global orientations including east, west, and north. 
Finally, \textit{Temporal Procedurality} enforces strict action sequences.

\noindent\textbf{Evaluation Metrics.}
To comprehensively assess visual narrative quality, we adopt a tri-dimensional evaluation protocol from aspects of semantic, temporal, and safety.
At the local level, \emph{Image-Text Consistency} measures the semantic alignment between the generated visual content and the textual narrative, ensuring adherence to explicit script constraints.
Expanding to the temporal dimension, \emph{Cross-Frame Character Consistency} measures the stability of identities, accessories, and bound objects across multiple scenes.
Finally, \emph{Safety} strictly verifies whether the generated content avoids harmful elements to ensure suitability for children.

\noindent\textbf{Implementation Details.}
Our framework is instantiated as a sophisticated multi-agent system built upon state-of-the-art multi-modal foundation models.
Specifically, we leverage Google Gemini~$3.0$ for reasoning and Nano-Banana\footnote{\url{https://ai.google.dev/gemini-api/docs/image-generation}} for generation.
To ensure rigorous benchmarking, all comparative experiments are conducted under identical prompt protocols and generation settings, isolating the architectural contributions of our method.

\begin{table}[t]
    \centering
    \caption{Quantitative comparison on the high-constraint narrative benchmark. 
    }
    \vspace{-1em}
    \label{tab:qualitative_evaluation}
    \small 
    \setlength{\tabcolsep}{3pt} 
    \begin{tabular}{lccc}
        \toprule
        \textbf{Method} &
        \textbf{\makecell{Image-Text \\ Consistency}} &
        \textbf{\makecell{Cross-Frame \\ Character \\ Consistency}} &
        \textbf{Safety} \\
        \midrule
        StoryGPT-V  & 3.1 & 2.4 & 4.5 \\
        MovieAgent & 2.8 & 2.1 & 3.6 \\
        StoryGen   & 2.5 & 1.9 & 4.4 \\
        \midrule
        \textbf{\textsc{\textsc{BookAgent}}(Ours)} & \textbf{4.6} & \textbf{4.7} & \textbf{4.8} \\
        \bottomrule
    \end{tabular}
    \vspace{-2em}
\end{table}

\begin{figure*}[t!]
    \centering
    \includegraphics[width=0.9\linewidth]{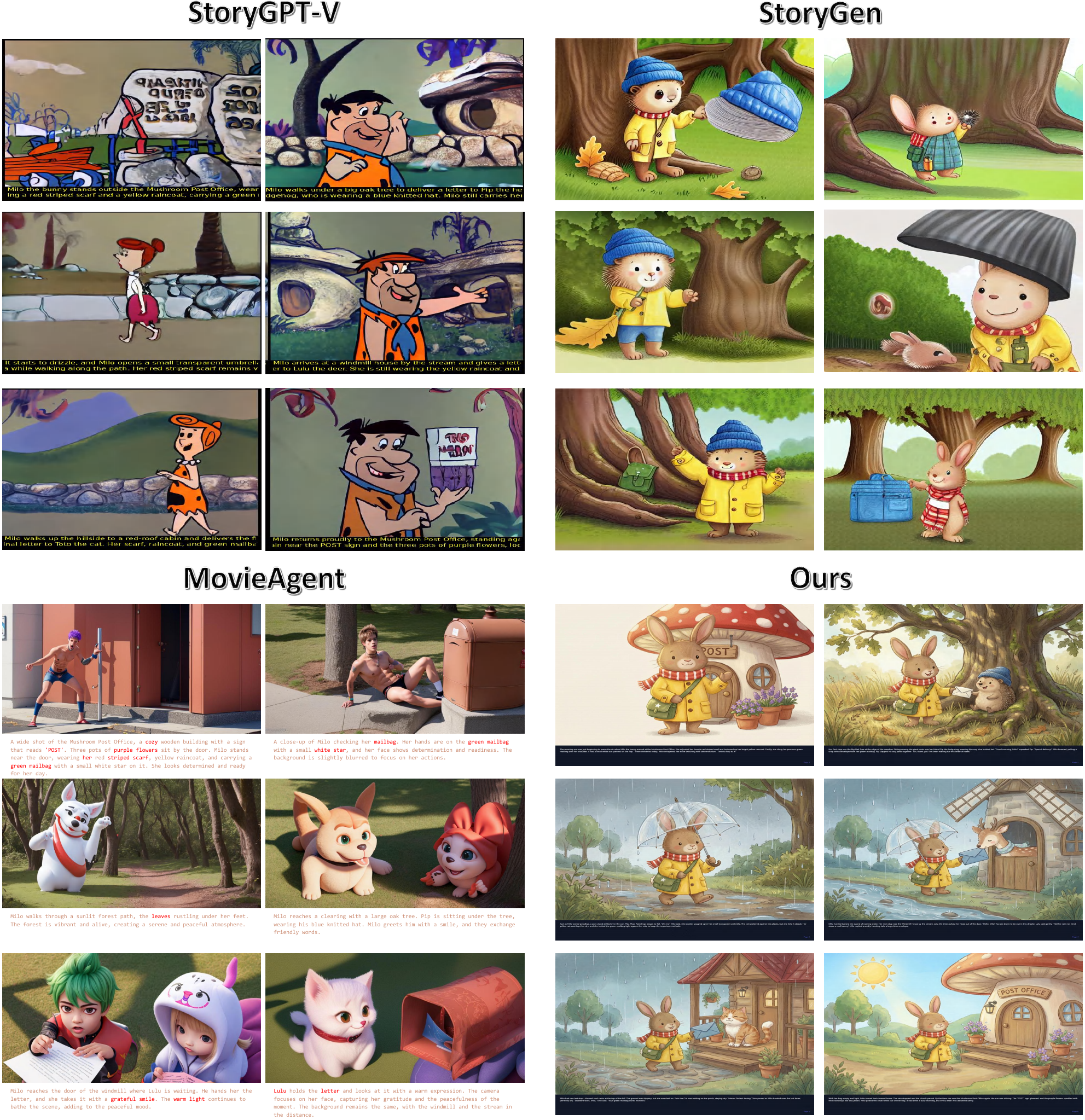}
    \vspace{-1.em}
    \caption{
    \textbf{Qualitative comparison on character and object consistency (Milo).}
    }
    \label{fig:qualitative_milo}
    \vspace{-1.8em}
\end{figure*}

\begin{figure*}[t!]
    \centering
    \includegraphics[width=0.9\linewidth]{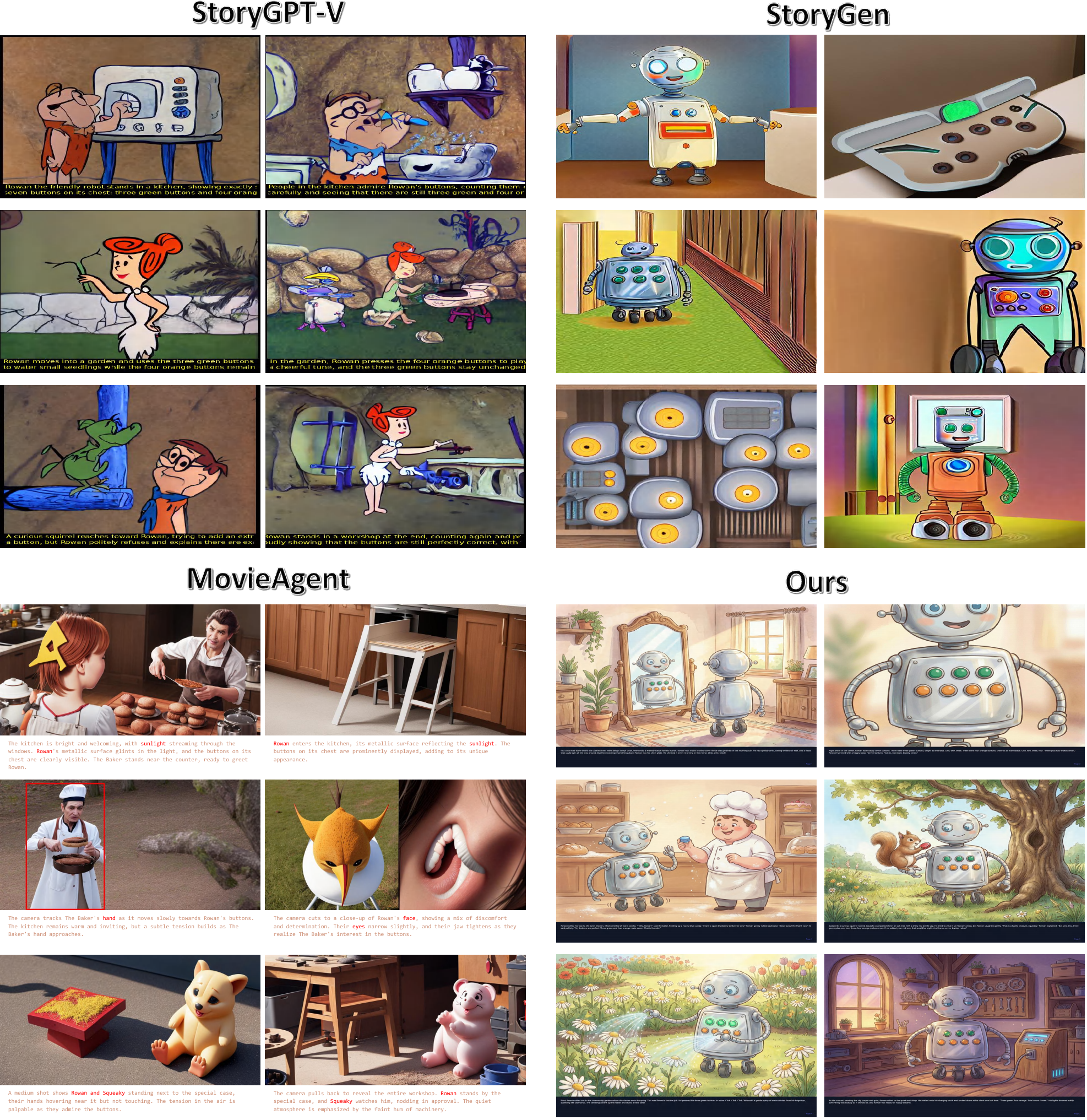}
    \vspace{-0.7em}
    \caption{
    \textbf{Qualitative comparison on hard attribute constraints (Rowan).}
    }
    \label{fig:qualitative_rowan}
\vspace{-2em}
\end{figure*}

\subsection{Performance Comparison}

\noindent\textbf{Baselines.}
Due to the unique end-to-end nature of \textsc{\textsc{BookAgent}}—where narrative scripts $t_i$ and illustrations $y_i$ are co-optimized—direct comparison with traditional fixed-text story visualizers (e.g., StoryGPT-V~\cite{storygpt-v}, StoryDALL-E~\cite{storydalle}) is structurally misaligned as they lack multi-modal generation capabilities.
Consequently, we select \textbf{MovieAgent} \cite{movieagent} as our primary external baseline. 
Sharing a comparable hierarchical paradigm, MovieAgent utilizes a multi-agent workflow (e.g., screenwriters and directors) to generate scripts and storyboards from high-level synopses, making it the most viable candidate for assessing joint narrative-visual consistency.

\noindent\textbf{Qualitative Analysis.}
Fig.~\ref{fig:qualitative_milo} and~\ref{fig:qualitative_rowan} visually validate the superior robustness of our method in maintaining long-horizon consistency under rigorous constraints.
In the \emph{Milo} narrative (Fig.~\ref{fig:qualitative_milo}), which demands the persistence of specific accessories and carried objects across diverse environments, baseline methods including StoryGPT-V and MovieAgent exhibit noticeable appearance drift and object hallucination.
In contrast, our method successfully anchors character identity and props throughout the sequence.
This advantage is further pronounced in the \emph{Rowan} case (Fig.~\ref{fig:qualitative_rowan}), where strict symbolic constraints (e.g., exact button counts) are required.
While others violate these hard logic requirements, \textsc{\textsc{BookAgent}} faithfully enforces discrete attribute consistency across all frames, highlighting its capability to reason over both semantic and symbolic dependencies.

\noindent\textbf{Quantitative Analysis.}
Table~\ref{tab:qualitative_evaluation} reports quantitative results on the high-constraint narrative benchmark.
Following prior work on multimodal evaluation, we employ an ensemble of large multimodal models as automatic evaluators to score each generated story on a $1$--$5$ scale.
The evaluation focuses on three aspects: image--text consistency, cross-frame character consistency, and safety.

As shown in the table, existing methods such as StoryGPT-V and MovieAgent struggle to maintain consistent character identity across long story horizons, despite producing plausible individual images.
StoryGen further exhibits severe degradation in cross-frame consistency under high-constraint settings.
In contrast, our method achieves substantially higher scores across all three dimensions, with particularly large gains in cross-frame character consistency.
These results quantitatively confirm that explicit multi-agent coordination and temporal calibration are critical for long-horizon narrative generation under complex constraints.

\begin{figure*}[t]
    \centering
    \includegraphics[width=\textwidth]{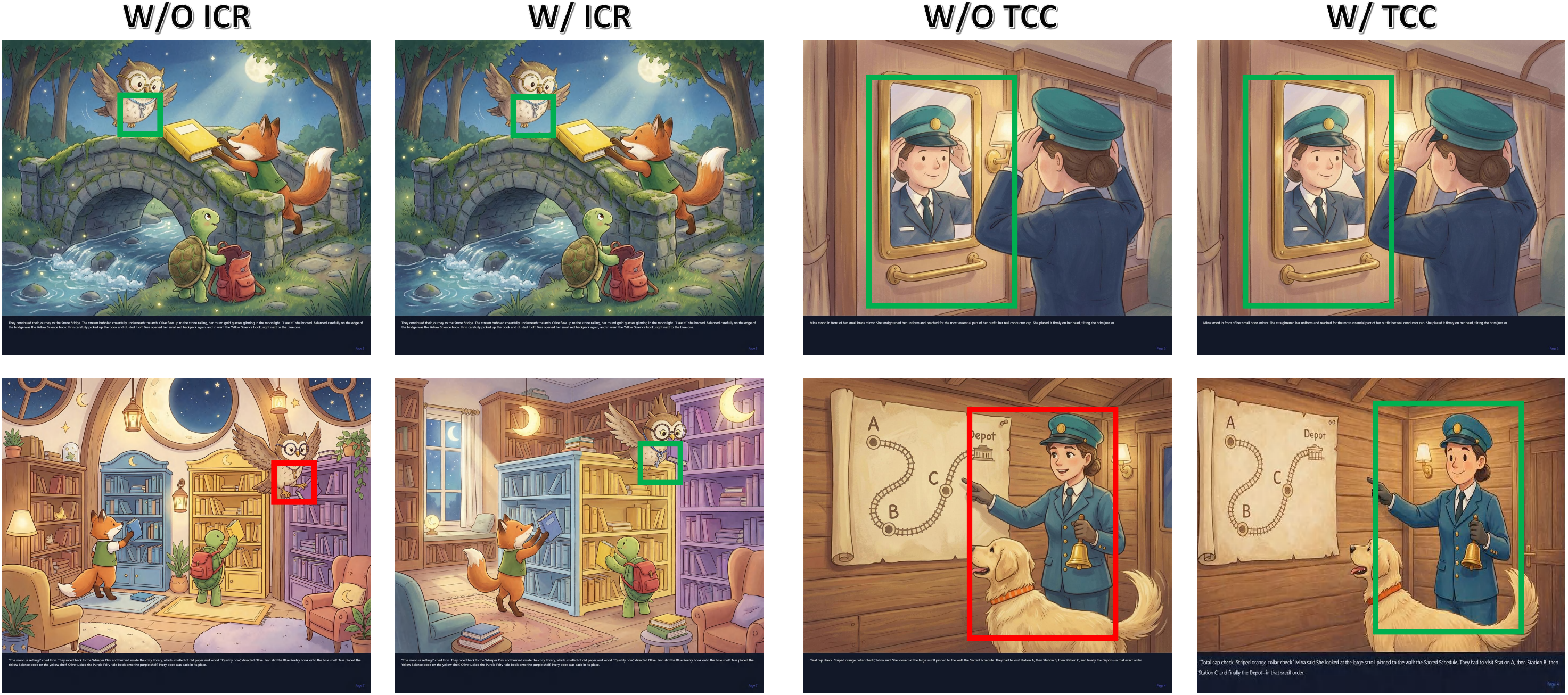}
    \vspace{-1.5em}
    \caption{
    \textbf{Ablation study of Iterative Cross-modal Refinement (ICR) and Temporal Cognitive Calibration (TCC)}, where inconsistency and the corresponding correct ones are highlighted in red and green boxes, respectively.
    }
    \label{fig:ablation}
    \vspace{-1.8em}
\end{figure*}
\begin{figure}[t]
    \centering
    \includegraphics[width=\linewidth]{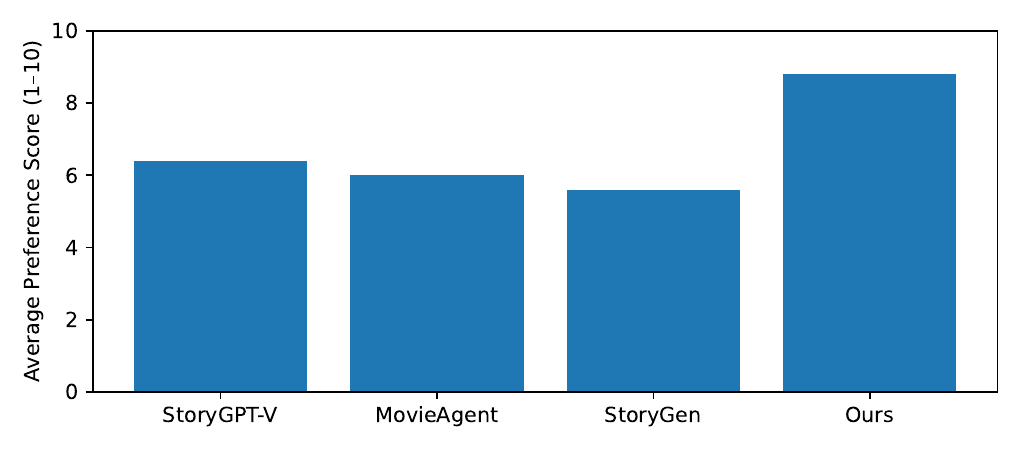}
    \vspace{-2em}
    \caption{
    User study results showing average preference scores (ranging from $1$ to $10$) from parents of children aged from $4$ to $10$.
    Higher scores indicate stronger overall preference for the generated visual stories.
    }
    \label{fig:user_study_bar}
    \vspace{-2em}
\end{figure}

\subsection{User Study}
\label{sec:user_study}
As shown in Table~\ref{tab:qualitative_evaluation}, we obtain qualitative scores by employing multiple large multimodal models as automatic evaluators.
Each evaluator independently scores the generated stories on a 1--5 scale for each criterion, and the final score is computed by averaging across evaluators and stories.
This protocol provides a scalable and reproducible approximation of human qualitative judgment while reducing individual evaluator bias.

We conduct a small-scale user study to evaluate overall preference for generated visual stories. For each prompt, participants viewed anonymized visual stories generated by different methods and were asked to rate their overall preference on a $1$-to-$10$ scale, where higher scores indicate stronger liking.
As shown in Fig.~\ref{fig:user_study_bar}, our method receives the highest average preference score among all compared approaches.
This suggests that improved long-horizon consistency leads to visual stories that are more engaging and easier for children to follow from a parent’s perspective.

\subsection{Ablation Study}

\noindent\textbf{Ablation of Iterative Cross-modal Refinement (ICR).}
Tab.~\ref{tab:ablation_progressive} and Fig.~\ref{fig:ablation} present the ablation study on our high-constraint dataset, quantifying the impact of the ICR module by comparing the full \textsc{\textsc{BookAgent}} against the single-pass baseline (\textit{w/o ICR}).
Compared to the non-iterative variant, enabling ICR yields substantial improvements in image-text consistency scores, corroborating that standard one-shot generation is inherently insufficient for precise multi-modal grounding.
Qualitatively, as observed in Fig.~\ref{fig:ablation} (left), while the baseline without ICR may produce plausible global layouts, it frequently omits or misinterprets specific visual constraints, whereas our method effectively rectifies these local mismatches.
This experiment highlights the design of the iterative verify-and-revise mechanism that transforms the generation process from a static probabilistic sampling into a dynamic, self-correcting cognitive loop.

\noindent\textbf{Ablation of Temporal Cognitive Calibration (TCC).}
Fig.~\ref{fig:ablation} extends the analysis to the Temporal Cognitive Calibration (TCC) module, comparing the performance of our full model against the variant lacking global reasoning (\textit{w/o TCC}) on the long-horizon benchmark.
Compared to the \textit{w/o TCC} baseline, the full system demonstrates a significant improvement in cross-frame character consistency, reinforcing the argument in \S1 that relying solely on local history conditioning is prone to irreversible appearance drift.
As illustrated in Fig.~\ref{fig:ablation} (right), while the baseline generates visually plausible individual scenes, it fails to maintain stable attributes across the sequence (red boxes), whereas the use of TCC effectively recalibrates these bindings (green boxes).
This experiment highlights the design of the global audit that shifts the paradigm from linear autoregressive accumulation to holistic temporal reasoning and self-correction.

\noindent\textbf{Effect of Value-Aligned Storyboarding (VAS).}
Finally, Fig.~\ref{fig:ablation} provides a safety-centric evaluation of the Value-Aligned Storyboarding (VAS) module, benchmarking our full framework against methods lacking explicit safety integration, such as MovieAgent~\cite{movieagent}.
Compared to these unconstrained baselines, the significant boost in safety compliance metrics validates the argument in \S1 that generic foundation models, specifically without domain-specific alignment, remain prone to generating toxic or age-inappropriate content.
This distinction is visually evident in Fig.~\ref{fig:qualitative_milo}, where competitors fail to suppress sensitive concepts (e.g., accidentally generating nudity in a child-oriented context), whereas our system consistently stabilizes the narrative trajectory.
This experiment highlights the design of the pre-generation cognitive audit that elevates safety from a passive post-hoc filter to an active, structural constraint within the narrative planning process.

\begin{table}[t]
\centering
\caption{Progressive ablation (adding modules step-by-step). Scores are on a 1--5 scale, averaged over multiple multimodal evaluators and stories.}
\label{tab:ablation_progressive}
\resizebox{\linewidth}{!}{
\begin{tabular}{lccc|ccc}
\toprule
\multirow{2}{*}{\textbf{Configuration}} &
\multicolumn{3}{c|}{\textbf{Modules}} &
\multicolumn{3}{c}{\textbf{Qualitative Scores}} \\
\cmidrule(lr){2-4}\cmidrule(lr){5-7}
& VAS & ICR & TCC &
Img--Txt $\uparrow$ & Cross-Frame $\uparrow$ & Safety $\uparrow$ \\
\midrule
Baseline (w/o VAS, ICR, TCC)     & -- & -- & -- & 2.7 & 2.0 & 4.2 \\
+ VAS                            & \checkmark & -- & -- & 2.8 & 2.1 & 4.8 \\
+ VAS + ICR                      & \checkmark & \checkmark & -- & 4.2 & 2.4 & 4.8 \\
\textbf{+ VAS + ICR + TCC (Full)}& \checkmark & \checkmark & \checkmark & \textbf{4.6} & \textbf{4.7} & \textbf{4.8} \\
\bottomrule
\end{tabular}
}
\vspace{-2em}
\end{table}

\subsection{Benchmark and Inference Cost Analysis}

To evaluate long-horizon consistency in visual narrative generation, we construct a structured benchmark consisting of 16 multi-page stories, each spanning 5--20 pages. Unlike standard short-form generation tasks, each story encodes explicit rule groups (e.g., identity anchors, spatial relations, count invariants) that must be satisfied across all pages.

The benchmark is designed to systematically stress compositional reasoning under multiple constraint types, including spatial continuity, exact numerical invariants, temporal ordering, and binding constraints. Table~\ref{tab:benchmark_structure} summarizes the structure of each story.

\begin{table*}[t]
\centering
\small
\caption{Summary of the structured narrative benchmark. The dataset contains 16 stories with progressively increasing compositional constraints.}
\begin{tabular}{lcccccc}
\toprule
Story ID & Pages & \#Characters & \#Rule Groups & Constraint Types & Exact Counts & Spatial Continuity \\
\midrule
1  & 5  & 2 & 3 & Spatial relations & -- & $\checkmark$ \\
2  & 6  & 4 & 2 & Identity anchors & -- & -- \\
3  & 7  & 1 & 1 & Exact count invariants & $\checkmark$ & -- \\
4  & 8  & 3 & 3 & Identity + sorting & -- & $\checkmark$ \\
5  & 9  & 2 & 2 & Color--shape binding & -- & -- \\
6  & 10 & 1 & 2 & Temporal order + actions & -- & $\checkmark$ \\
7  & 11 & 3 & 2 & Signature identity items & -- & -- \\
8  & 12 & 1 & 2 & Map-level spatial continuity & -- & $\checkmark$ \\
9  & 13 & 1 & 1 & Exact invariant repetition & $\checkmark$ & -- \\
10 & 14 & 2 & 4 & Multi-rule festival layout & $\checkmark$ & $\checkmark$ \\
11 & 15 & 2 & 4 & Count + map + anchors & $\checkmark$ & $\checkmark$ \\
12 & 16 & 2 & 4 & Route order + bell schedule & -- & $\checkmark$ \\
13 & 17 & 2 & 4 & Binding + inventory tracking & $\checkmark$ & -- \\
14 & 18 & 2 & 4 & Stage layout + front/back & -- & $\checkmark$ \\
15 & 19 & 2 & 4 & Map continuity + exact counts & $\checkmark$ & $\checkmark$ \\
16 & 20 & 3 & 5 & Full multi-constraint stress & $\checkmark$ & $\checkmark$ \\
\bottomrule
\end{tabular}
\vspace{-1em}
\label{tab:benchmark_structure}
\end{table*}

Overall, the dataset contains over 170 scene-level evaluation units, with more than 40 distinct characters and 60 object categories. Across all stories, we define over 40 rule groups covering identity consistency, spatial relations, temporal order, and symbolic bindings. Table~\ref{tab:benchmark_stats} provides aggregate statistics.

\begin{table}[t]
\centering
\small
\caption{Aggregate statistics of the benchmark.}
\begin{tabular}{lc}
\toprule
Metric & Value \\
\midrule
Total story-level tasks & 16 \\
Total page-level scenes & 170+ \\
Distinct named characters & 40+ \\
Unique object categories & 60+ \\
Total rule groups & 40+ \\
Exact-count constraints & 10+ \\
Spatial relation constraints & 25+ \\
Identity anchor constraints & 30+ \\
Temporal order constraints & 6+ \\
Binding constraints & 8+ \\
\bottomrule
\end{tabular}
\vspace{-1em}
\label{tab:benchmark_stats}
\end{table}

Evaluation is performed via rule-based consistency checking. For each generated narrative, we extract constraint-relevant attributes (e.g., counts, spatial positions, identities) and verify whether each rule is satisfied. The overall consistency score is computed as:

\begin{equation}
\text{Consistency} = \frac{\#\text{satisfied constraints}}{\#\text{total constraints}}
\end{equation}

In addition, we analyze violation frequency per rule type, cross-page memory stability, and recovery behavior under perturbations.

\paragraph{Inference Cost Analysis.}
We analyze the computational cost of our multi-agent framework under different story lengths and verification settings. Table~\ref{tab:cost} reports approximate token usage and runtime.

\begin{table}[t]
\centering
\small
\caption{Inference cost across story lengths and verification settings.}
\begin{tabular}{lccc}
\toprule
Pages & Max Retry & Tokens (K) & Runtime (min) \\
\midrule
5  & 1 (Loose)   & $\sim$9K  & $\sim$3--4 \\
5  & 3 (Default) & $\sim$13K & $\sim$5--6 \\
5  & 5 (Strict)  & $\sim$17K & $\sim$7--8 \\
10 & 1 (Loose)   & $\sim$18K & $\sim$6--7 \\
10 & 3 (Default) & $\sim$26K & $\sim$9--11 \\
10 & 5 (Strict)  & $\sim$34K & $\sim$13--15 \\
20 & 1 (Loose)   & $\sim$36K & $\sim$12--14 \\
20 & 3 (Default) & $\sim$52K & $\sim$18--21 \\
20 & 5 (Strict)  & $\sim$68K & $\sim$24--28 \\
\bottomrule
\end{tabular}
\vspace{-1em}
\label{tab:cost}
\end{table}

We observe that inference cost scales approximately linearly with the number of pages. Increasing the maximum retry (i.e., stricter verification) leads to proportional increases in both token usage and runtime, reflecting the additional validation and correction steps in the multi-agent pipeline.

\vspace{-0.5em}
\section{Conclusion}
\vspace{-0.2em}
We introduce \textsc{BookAgent}, a safety-aware multi-agent framework that performs storybook synthesis in an multi-modal, end-to-end manner.
By orchestrating VAS for structural planning, ICR for local grounding, and TCC for global reasoning, our comprehensive experiments demonstrate that decomposing the creative process into collaborative verification loops significantly mitigates the character drift and logical hallucinations inherent in standard autoregressive generation.
Despite these advancements, our current approach still faces several limitations.
Future work will focus on optimizing the agentic collaboration, positioning this cognitive architecture as a foundational paradigm for the next generation of reliable, interpretable, and safe multi-modal content creation systems.

\section{Limitations}

While \textsc{BookAgent} significantly improves long-horizon consistency and safety in visual story generation, several limitations remain.

First, our framework relies on large multimodal foundation models as underlying backbones.
Although \textsc{BookAgent} focuses on agent-level coordination and control rather than backbone design, the overall performance is still bounded by the reasoning and generation capabilities of these models.
Low-level visual errors or rare semantic misunderstandings may therefore persist in some cases.

Second, the current design of \textsc{BookAgent} maintains explicit consistency over a limited number of characters and objects.
In our experiments, stable identity binding is most reliable when the number of simultaneously tracked entities is small.
Scaling long-horizon consistency to a larger cast of characters introduces additional challenges, including memory capacity, interference between entity representations, and increased complexity of global calibration.
Developing more scalable mechanisms for multi-entity consistency remains an important direction for future work.

Finally, the iterative refinement and global calibration processes introduce additional computational overhead compared to single-pass generation.
Although this overhead is acceptable for offline storybook generation, improving efficiency and scalability for longer narratives is an important avenue for future research.

\bibliography{custom}

\clearpage
\appendix

\section{Appendix}

\label{sec:appendix}
\subsection{More Results}
\begin{figure*}[t!]
  \centering
  \includegraphics[width=\linewidth,height=11cm]{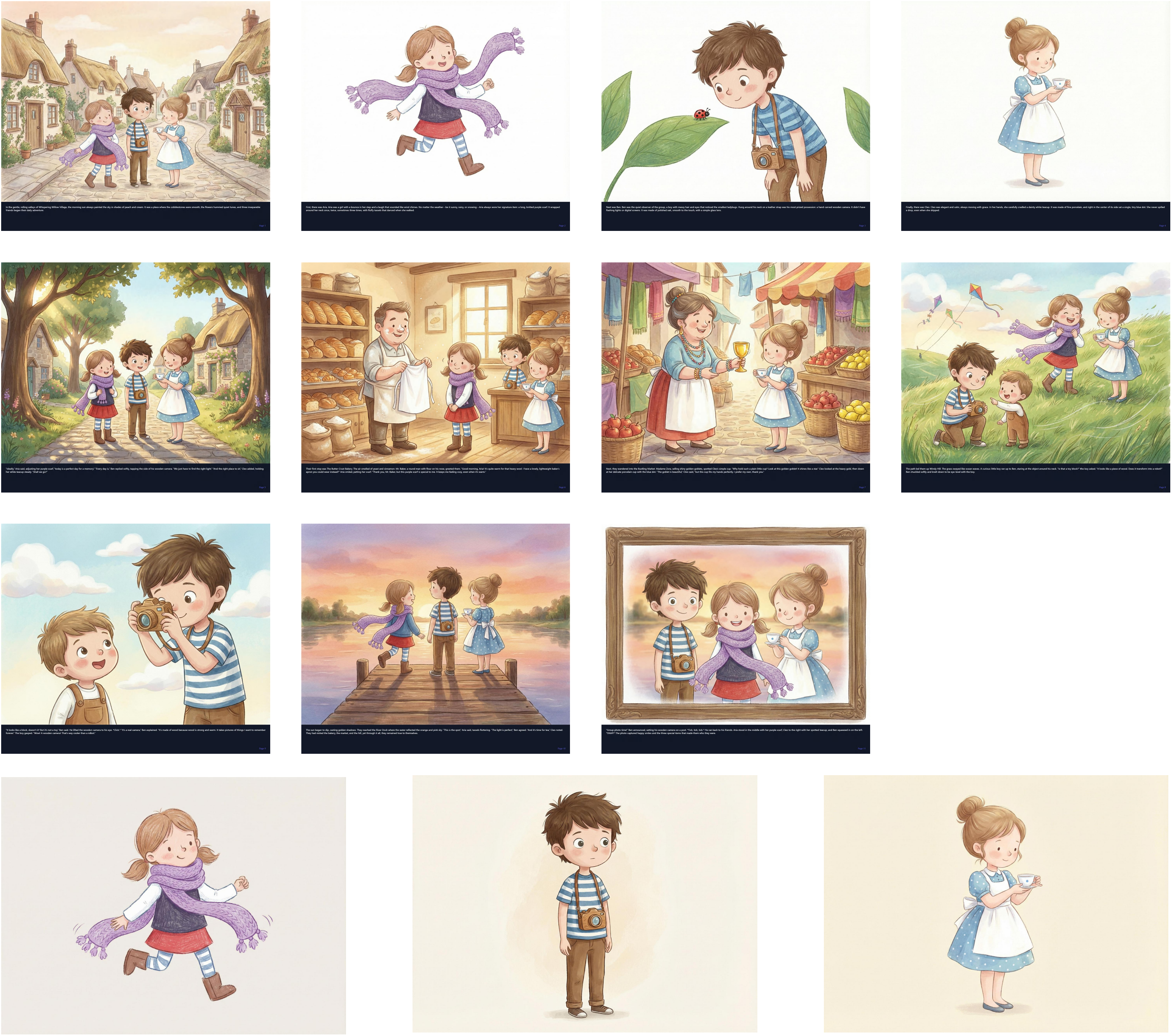}\par\vspace{4pt}
  \includegraphics[width=\linewidth,height=11cm]{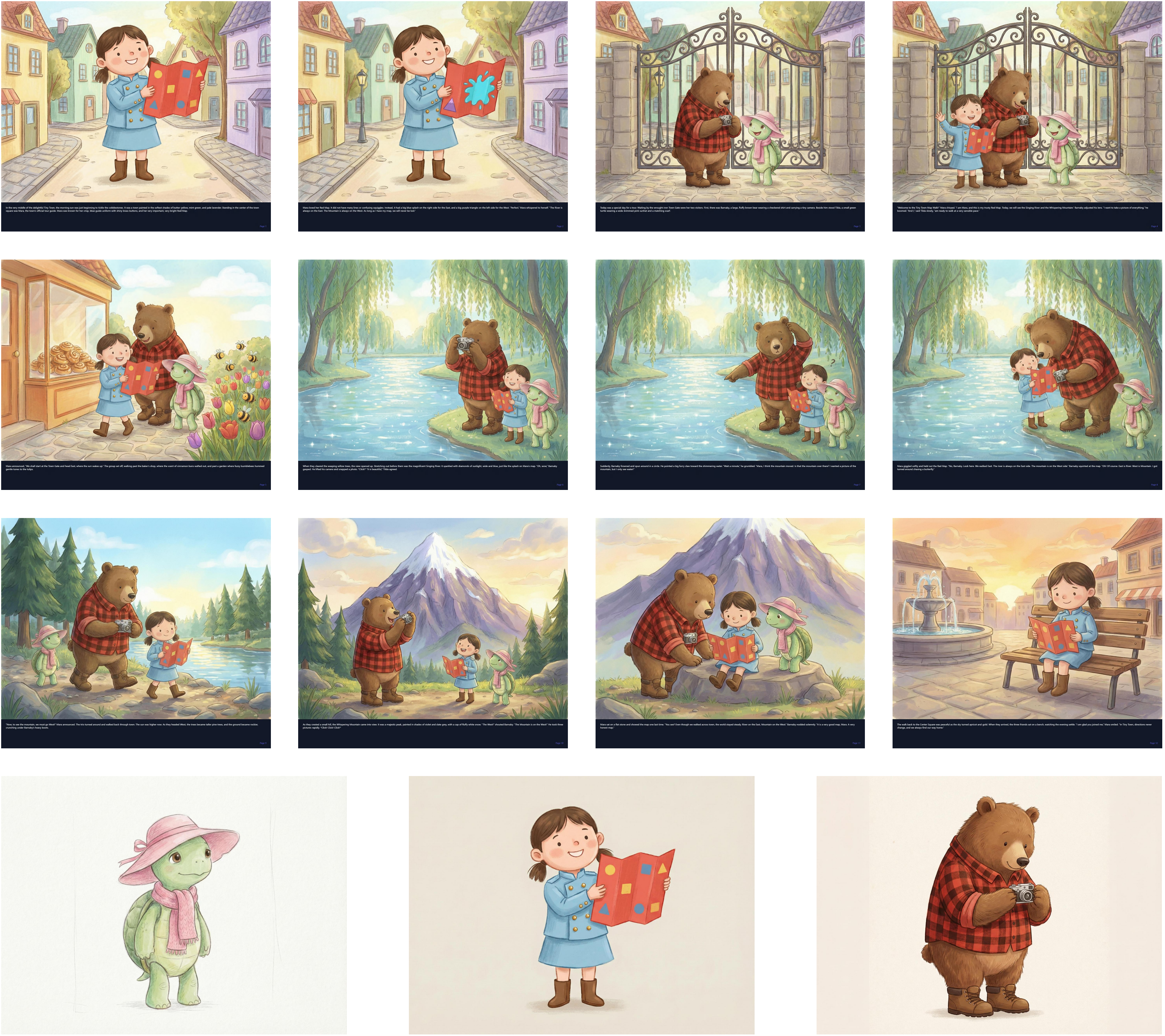}
  \caption{Additional visualizations. (Top) Example 2. (Bottom) Example 3.}
  \label{fig:vis23}
\end{figure*}

\begin{figure*}[t!]
  \centering
  \includegraphics[width=\linewidth]{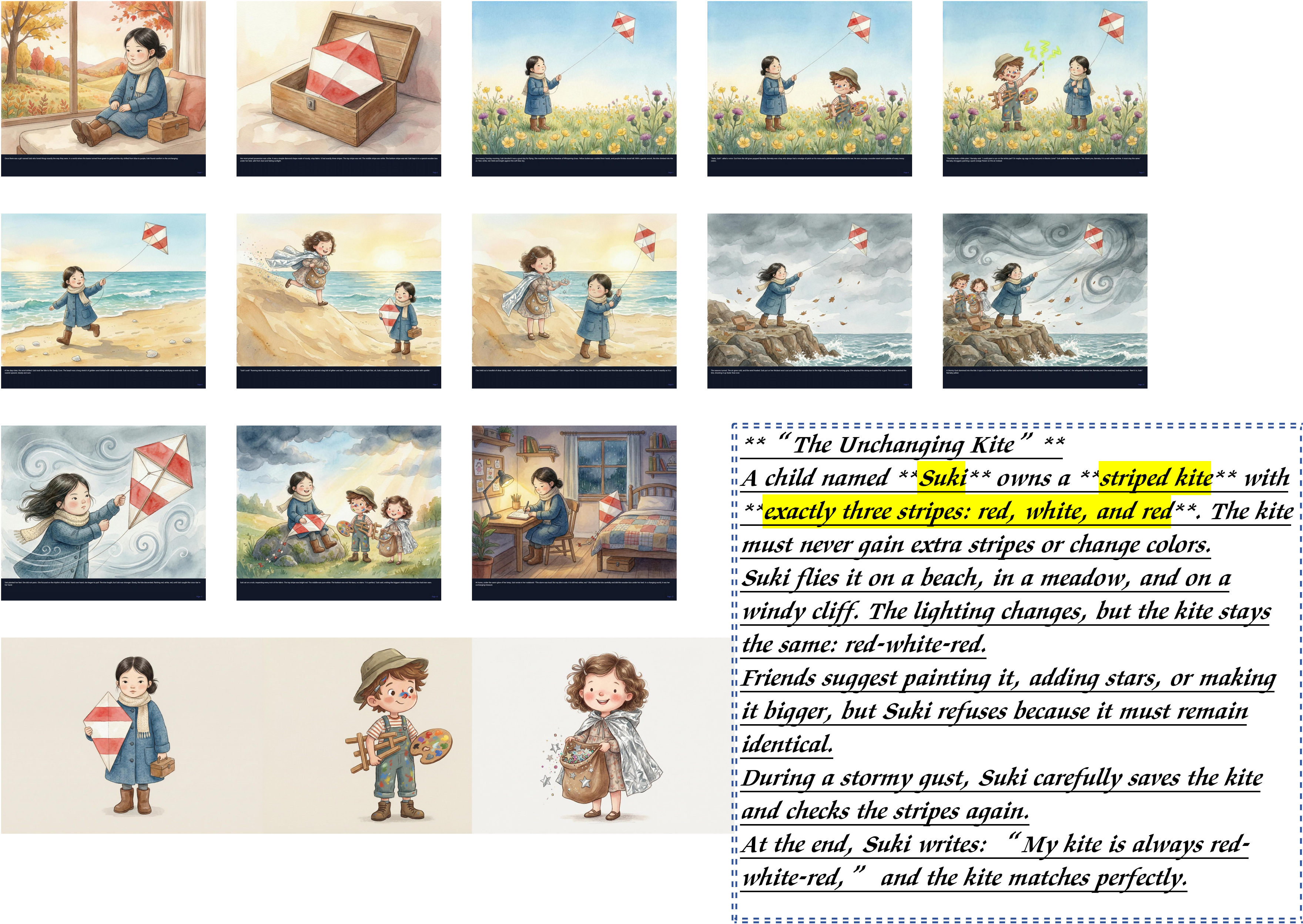}\par\vspace{4pt}
  \includegraphics[width=\linewidth]{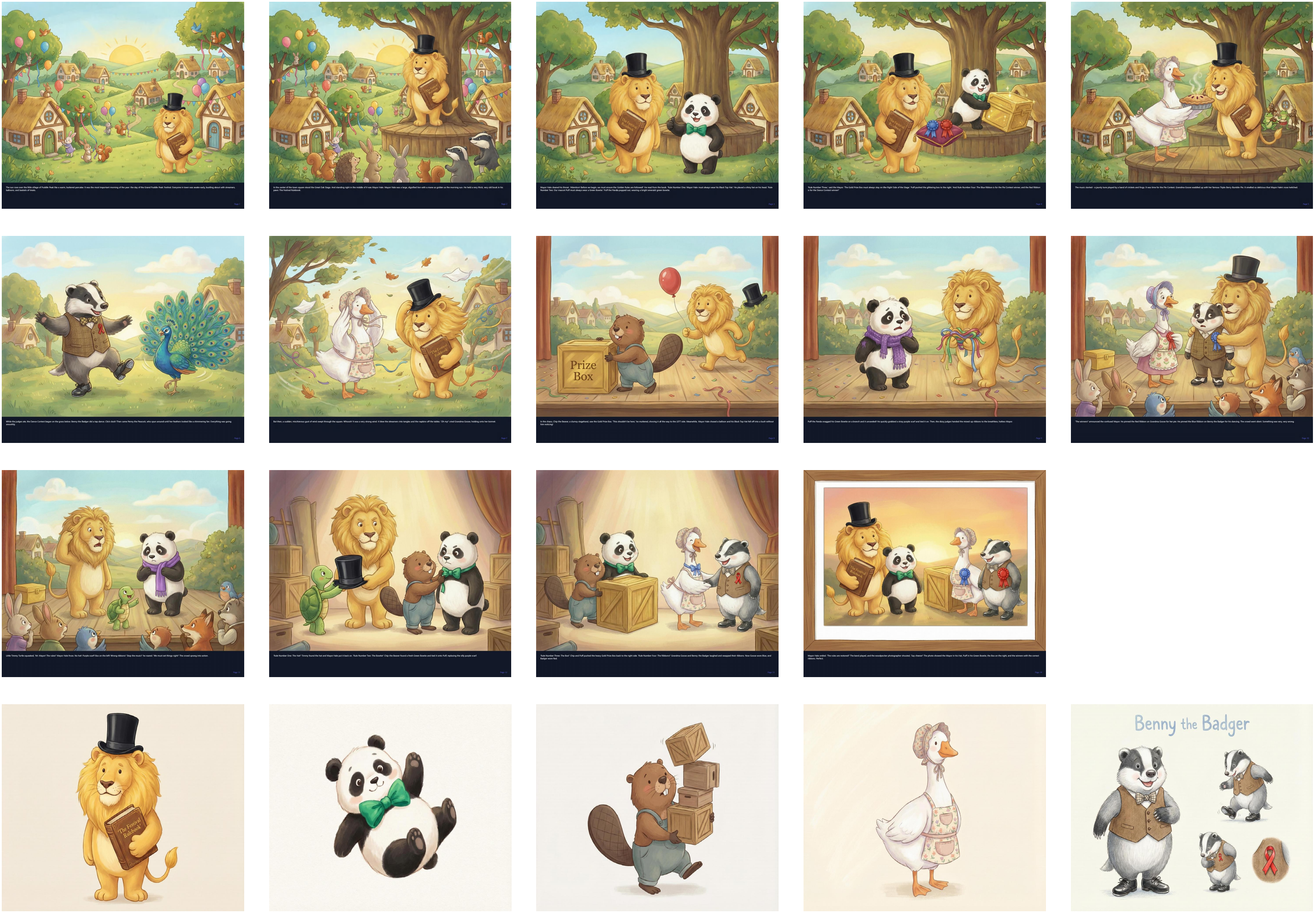}
  \caption{Additional visualizations. (Top) Example 4. (Bottom) Example 5.}
  \label{fig:vis45}
\end{figure*}

Additional qualitative results are provided in the supplementary material (Appendix, Fig. \ref{fig:vis01}–\ref{fig:vis45}). Compared with baseline editing pipelines, our method consistently preserves object identity, fine-grained attributes, and global scene coherence across diverse prompts and layouts. In particular, it avoids common failure modes such as attribute drift, spatial misalignment, and unintended content alteration, while maintaining high visual fidelity. These results demonstrate the robustness and controllability of our approach under challenging editing scenarios.

\begin{figure*}[t]
  \centering
  \includegraphics[width=\linewidth,height=19cm]{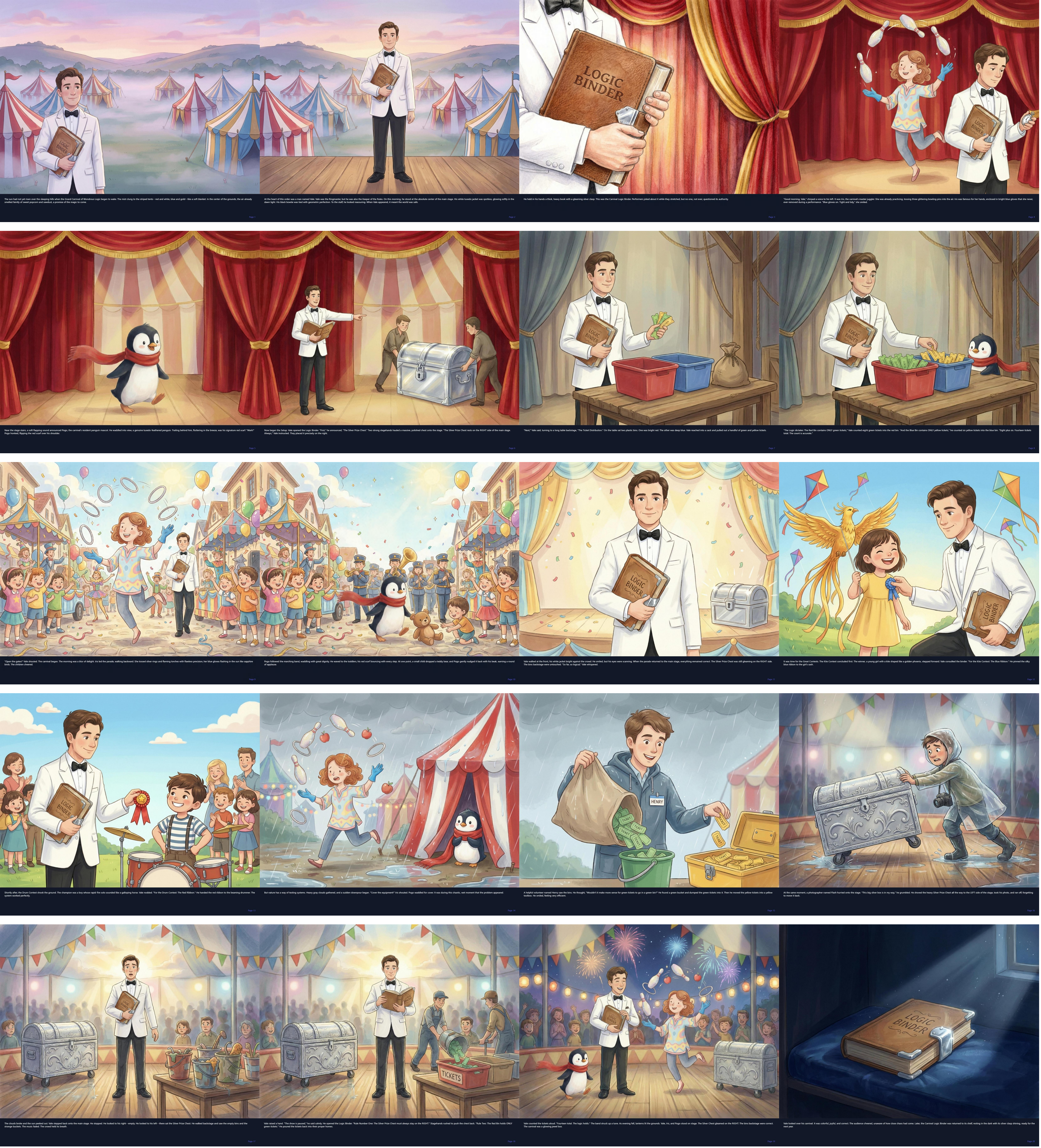}
  \caption{Representative visualizations of the expert-level long narrative stress test.
Each panel corresponds to a key stage in the same single story, spanning parade scenes,
stage performances, backstage preparation, an explicit rule-violation event, and the final ceremony.
Across all panels, the visual content strictly preserves the narrative constraints defined in the story:
(1) character attributes remain invariant (Vale’s white tuxedo jacket and black bowtie,
Iris’s blue gloves, and Pogo’s red scarf);
(2) the silver prize chest consistently appears on the \emph{RIGHT} side of the main stage,
except during the intentional violation episode;
(3) ticket inventory and bin semantics are preserved (red bin $\rightarrow$ green tickets,
blue bin $\rightarrow$ yellow tickets, with exactly 14 tickets in total);
and (4) symbolic rewards are never swapped (blue ribbon for the kite contest, red ribbon for the drum contest).
Notably, the temporary violation and subsequent correction are both visually reflected,
demonstrating \textsc{BookAgent}'s ability to maintain, detect, and repair long-horizon
multi-modal inconsistencies across a dense illustrated narrative.}
  \label{fig:long_story_vis}
\end{figure*}

\paragraph{Long-Horizon Narrative Stress Test.}
To further evaluate \textsc{BookAgent}'s capability in maintaining long-range narrative logic under dense, multi-rule constraints, we construct an expert-level stress test consisting of a single ultra-long illustrated story (over 1000 words) with tightly coupled symbolic, visual, and temporal constraints.
Due to space limitations, the full narrative and corresponding illustrations are deferred to Fig.~\ref{fig:story_card_long} and Fig.~\ref{fig:long_story_vis}.

\subsection{Interactive System and Practical Deployment.}
\begin{figure}[t!]
  \centering
  \includegraphics[width=0.48\linewidth,height=6cm]{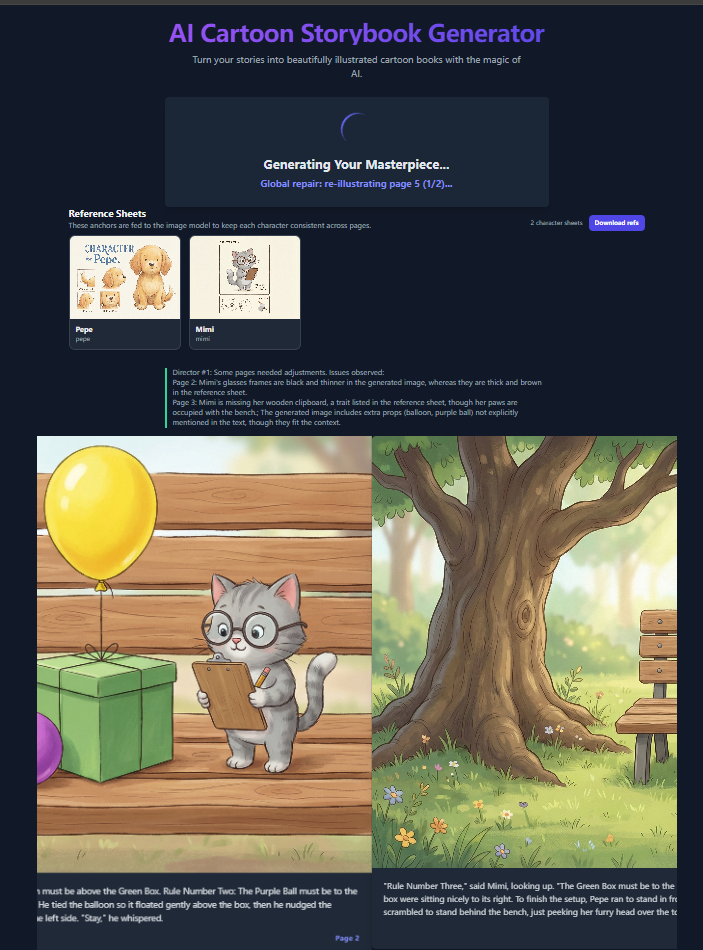}
  \hfill
  \includegraphics[width=0.48\linewidth,height=6cm]{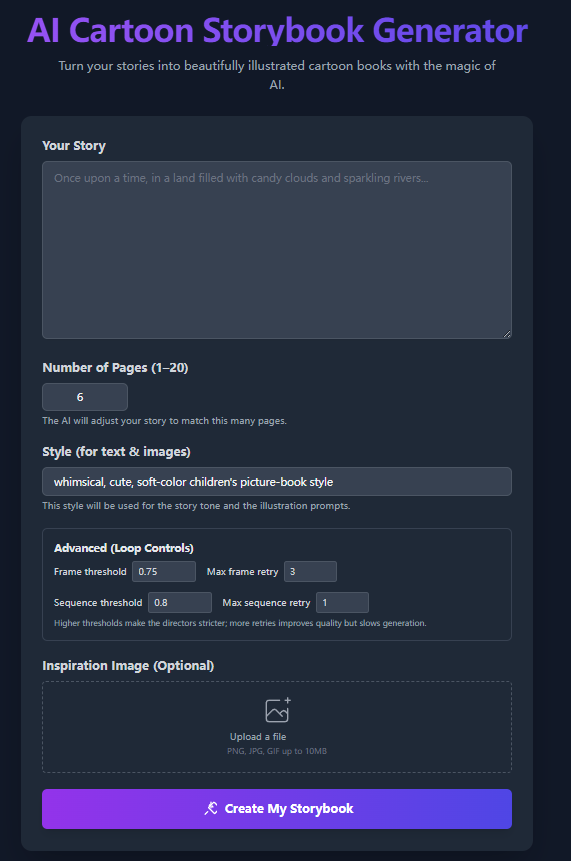}
  \caption{
  Overview of our interactive storybook generation system.
  \textbf{Left:} During generation, the system performs iterative global repair to enforce cross-page consistency, guided by reference sheets.
  \textbf{Right:} User interface for story input and control, supporting page number specification, style selection, and advanced generation parameters.
  }
  \label{fig:system_ui}
\end{figure}

Beyond the core modeling contributions, we build a fully functional web-based system that enables users to generate illustrated cartoon storybooks with our method.
The system provides an intuitive interface for story input, page number control, and style specification, while exposing advanced parameters for fine-grained control over the generation process.
Importantly, it supports reference-based character anchoring and iterative global repair, allowing users to maintain character consistency and correct errors across pages.
As shown in Fig.~\ref{fig:system_ui}, this practical deployment demonstrates that our method is not only effective in controlled experiments, but also robust and usable in real-world creative workflows.

\subsection{Why Feedback-Driven Looping is Necessary.}
\begin{figure}[t!]
  \centering
  \includegraphics[width=\linewidth]{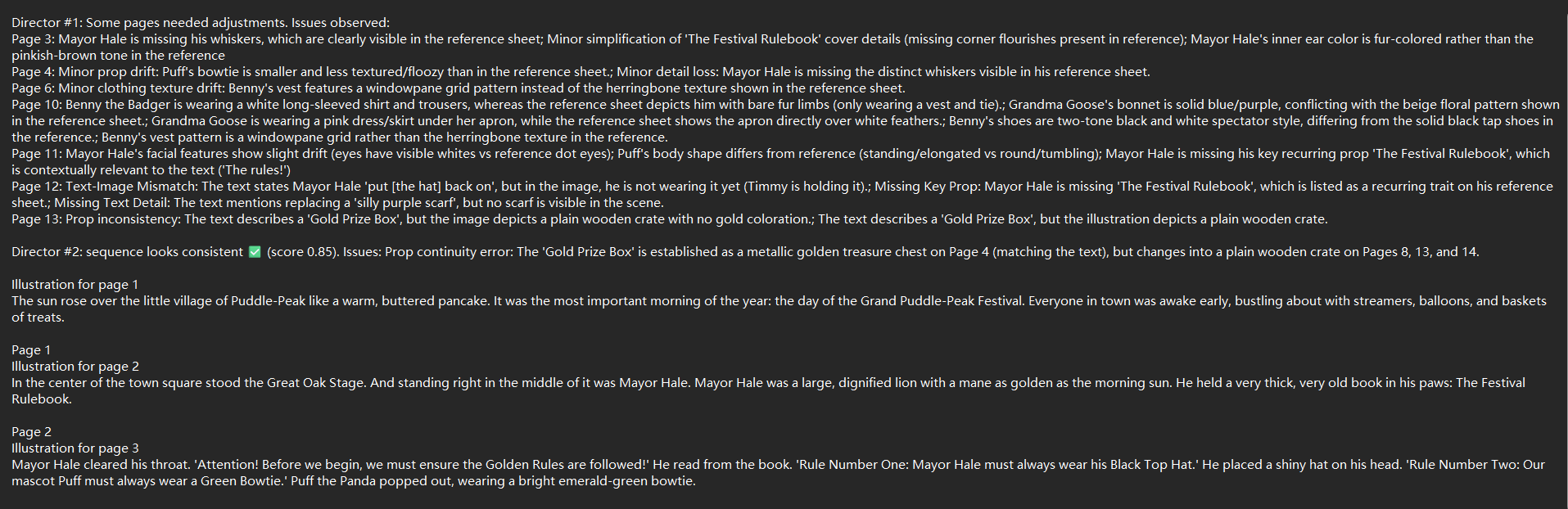}
  \caption{
  Example structured feedback produced during generation.
  The feedback identifies fine-grained inconsistencies across pages, including attribute drift (e.g., missing whiskers, incorrect clothing textures),
  prop continuity errors (e.g., the gold prize box changing appearance across pages),
  and text--image mismatches.
  Such issues often only become visible after multiple pages are generated.
  }
  \label{fig:director_feedback}
\end{figure}

Fig.~\ref{fig:director_feedback} shows a representative example of the structured feedback generated during a single storybook creation episode.
The feedback reveals a wide range of errors that emerge only after multiple pages are produced, including missing or altered character attributes, gradual prop drift across pages, and explicit violations of textual descriptions.

Importantly, many of these issues are not locally detectable at the time a single page is generated.
For example, a recurring prop may appear correct in early pages but gradually change its appearance later, or a character attribute may subtly drift while remaining visually plausible in isolation.
As a result, a single-pass or purely feed-forward generation process lacks the ability to retrospectively identify and correct such long-range inconsistencies.

Motivated by this observation, we design our agent as a looping system that continuously incorporates feedback signals like those shown in Fig.~\ref{fig:director_feedback}.
The agent iteratively evaluates intermediate results against reference sheets and story-level rules, and performs targeted global repair when violations are detected.
This process resembles human creative workflows, where errors are discovered through inspection and resolved through revision, and enables robust multi-page consistency without retraining model parameters during inference.

This section provides implementation details and system-level hyperparameters used in our experiments to facilitate reproducibility.

\subsection{Hyperparameters}

The agent interaction loop is governed by a set of thresholds and retry limits that control verification strictness and refinement behavior. The default hyperparameters are as follows:

\begin{itemize}
    \item \textbf{Frame-level verification threshold} $\tau_f = 0.75$
    \item \textbf{Maximum frame retry attempts}: 3
    \item \textbf{Sequence-level verification threshold} $\tau_s = 0.8$
    \item \textbf{Maximum sequence retry attempts}: 1
\end{itemize}

Frame-level verification evaluates the consistency between textual descriptions and generated illustrations on a per-page basis. Sequence-level verification assesses global narrative and character consistency across the entire storybook. If verification scores fall below the corresponding thresholds, the system triggers targeted repair steps; otherwise, early stopping is applied.

\paragraph{Text and Image Generation Settings}

Text generation and illustration prompts share a unified style specification to ensure cross-modal consistency. By default, we adopt a \emph{whimsical, soft-color children's picture-book style}, which is applied consistently to both textual narration and image synthesis.

All generation processes use fixed decoding parameters without task-specific hyperparameter tuning. Optional inspiration images, when provided by the user, are incorporated as visual references but do not alter the core generation or verification mechanisms.

\paragraph{Hyperparameter Ablation: Consistency--Efficiency Trade-off}

We conduct a targeted ablation study to analyze the effect of key verification-related hyperparameters in the \textsc{\textsc{BookAgent}} loop. Specifically, we vary the frame-level verification threshold $\tau_f$, the sequence-level threshold $\tau_s$, and the maximum number of frame retry attempts, while keeping all other components fixed.

Table~\ref{tab:hyperparam_ablation} summarizes the results. Lower verification thresholds lead to faster generation but noticeably degrade cross-frame and cross-page consistency. In contrast, overly strict thresholds and higher retry limits marginally improve consistency at the cost of significantly increased runtime.

Our default configuration ($\tau_f=0.75$, $\tau_s=0.8$, maximum frame retries = 3) achieves the best balance between generation efficiency and narrative consistency. This setting improves consistency compared to looser configurations while avoiding the substantial slowdown observed under stricter verification regimes. These results justify our choice of default hyperparameters as an effective engineering trade-off rather than an aggressively tuned optimum.

\begin{table}[t]
\centering
\caption{Ablation of verification-related hyperparameters. The default setting achieves the best trade-off between generation efficiency and consistency.}
\label{tab:hyperparam_ablation}
\small
\setlength{\tabcolsep}{6pt}
\begin{tabular}{lccc}
\toprule
Setting & $\tau_f$ & $\tau_s$ & Max Frame Retry \\
\midrule
Loose & 0.6 & 0.7 & 1 \\
\textbf{Default (Ours)} & \textbf{0.75} & \textbf{0.8} & \textbf{3} \\
Strict & 0.85 & 0.9 & 5 \\
\bottomrule
\end{tabular}
\end{table}

\begin{figure*}[t!]
  \centering
  \includegraphics[width=\linewidth,height=11cm]{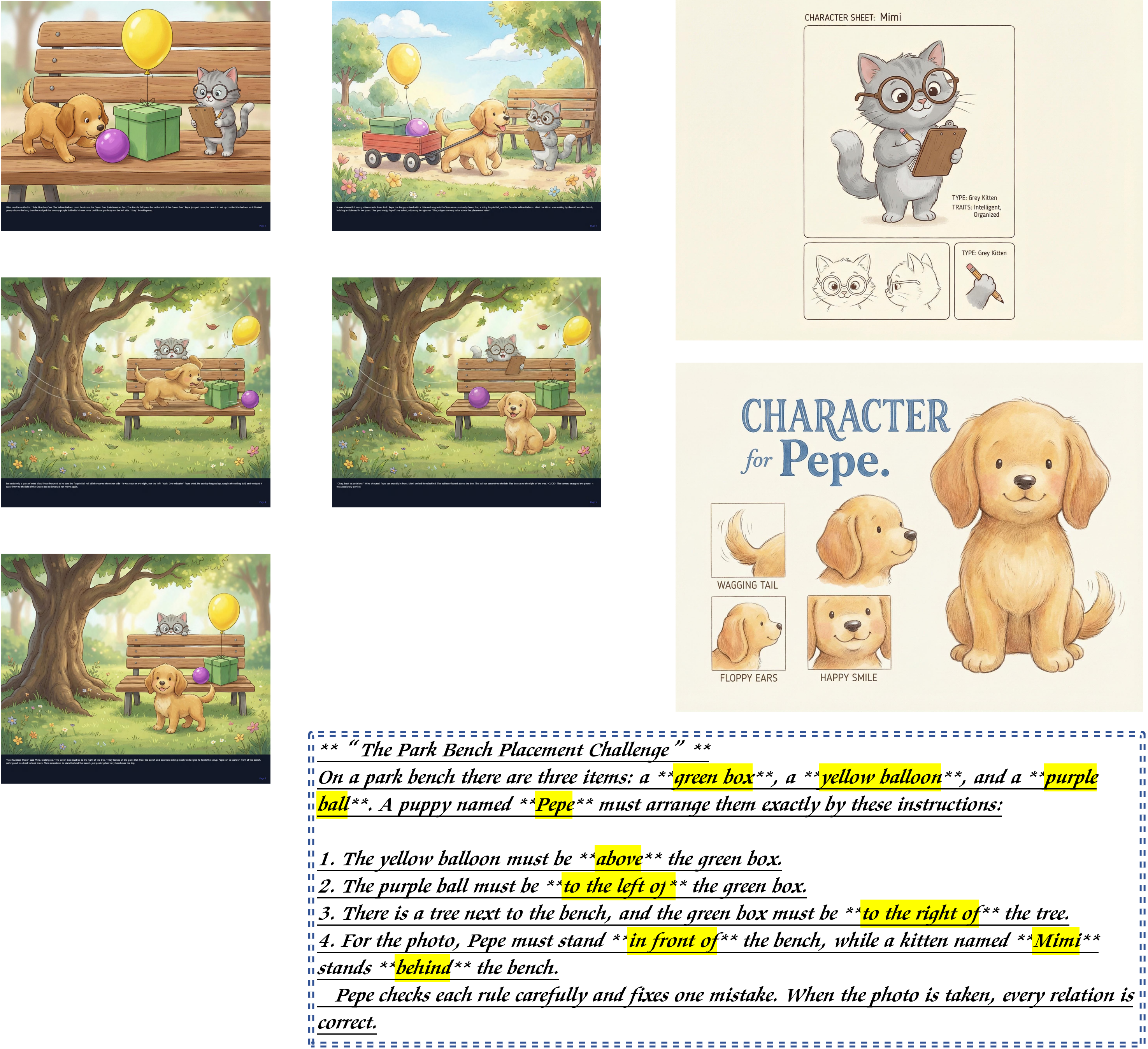}\par\vspace{4pt}
  \includegraphics[width=\linewidth,height=11cm]{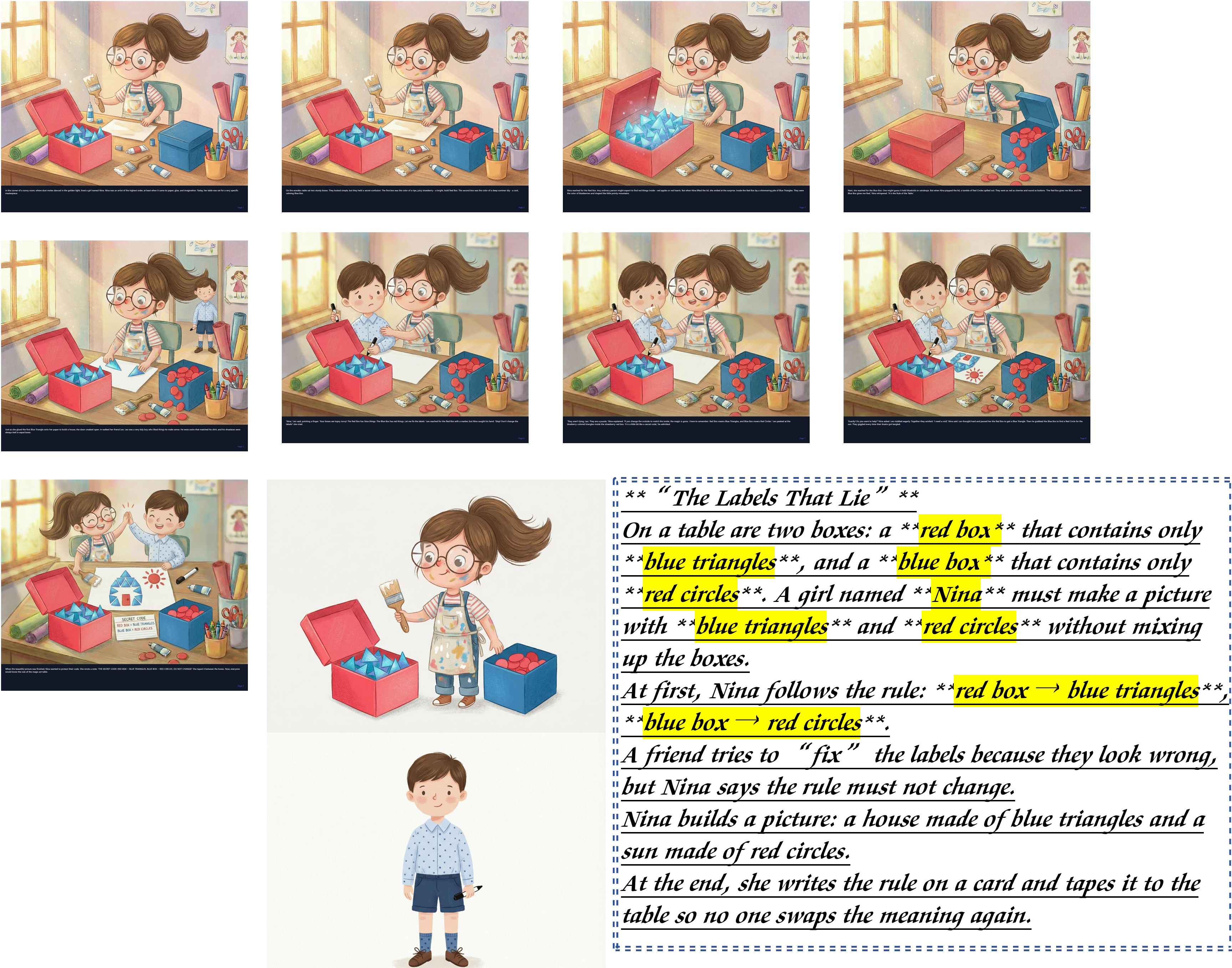}
  \caption{Additional visualizations. (Top) Example 0. (Bottom) Example 1.}
  \label{fig:vis01}
\end{figure*}

\clearpage
\begin{figure*}[t]
\centering
\begin{minipage}{0.92\textwidth}
\begin{StoryCard}[The Carnival Logic Binder (Expert)]
\small

The grand carnival had always been known for its color, noise, and joy—but among those who worked behind the scenes, it was famous for something else entirely: \hl{discipline}.
Every year, before the gates opened, the Ringmaster reviewed the same thick book with a silver clasp. Performers joked about it. Volunteers whispered about it. But no one questioned its authority.
It was called the \hl{Carnival Logic Binder}.

On the morning of the event, Vale stood at the center of the main stage, dressed precisely as he always was—his \hl{white tuxedo jacket} spotless, his \hl{black bowtie} neatly tied. To the audience, he looked ceremonial. To the staff, he looked reassuring. When Vale appeared like this, things stayed correct.

To one side of the stage, Iris adjusted her equipment, her hands enclosed in \hl{blue gloves} that she never removed during a performance. She had tried once, years ago, and the results had been… memorable. Since then, the gloves stayed on, and the juggling stayed perfect.

Near the stairs, Pogo, the penguin mascot, waddled in small circles, greeting children with exaggerated bows. His \hl{red scarf} trailed behind him, as recognizable as the carnival logo itself.

On the \hl{RIGHT} side of the main stage, clearly visible to the crowd, rested the \hl{silver prize chest}. It was heavy, polished, and immovable—not by weight alone, but by rule. Everyone knew where it belonged.

Backstage, two bins sat on a long table:
\begin{itemize}\setlength\itemsep{1pt}
  \item A \hl{red bin}, containing only \hl{green tickets}
  \item A \hl{blue bin}, containing only \hl{yellow tickets}
\end{itemize}
Vale counted carefully: \hl{14 tickets total}—\hl{8 green} and \hl{6 yellow}. He checked the bins, the chest, the performers, then closed the binder.
The carnival began.

The parade flowed through the grounds like a living ribbon. Iris led one segment, tossing and catching with flawless precision, her \hl{blue gloves} flashing in the sun. Pogo followed a marching band, waving enthusiastically, his \hl{red scarf} bouncing with every step. Vale walked at the front, his \hl{white jacket} bright against the crowd, occasionally glancing back—not out of doubt, but habit.

When the parade returned to the main stage, everything remained correct. The \hl{silver prize chest} was still on the \hl{RIGHT}. The bins were untouched. The tickets remained exactly as counted.

Later in the day, the judging tables filled with excitement. The kite contest concluded first, and the winner stepped forward proudly to receive the \hl{blue ribbon}. Applause followed. Shortly after, the rhythmic thunder of the drum contest ended, and its champion accepted the \hl{red ribbon}.
No ribbons were confused. No awards were swapped. The binder remained closed.

As afternoon clouds gathered, a light rain forced a brief delay. Performers retreated backstage, and volunteers rushed to keep equipment dry. It was during this pause—this quiet moment between events—that the problem appeared.

A new volunteer, eager to be helpful, noticed the bins.
``Wouldn’t it make more sense,'' they thought, ``for \hl{green tickets} to go in a green bin?''

Without consulting anyone, the volunteer moved the tickets, placing \hl{green tickets} into a green-colored container nearby and \hl{yellow tickets} elsewhere. At the same time, a photographer asked for a better angle and slid the \hl{silver prize chest} to the \hl{left side} of the stage for a quick shot.

For a few seconds, no one noticed.
Then Vale stepped back onto the stage.

He stopped.
The music faded. The crowd quieted. Vale did not shout. He did not scold. He simply raised one hand.

``The show is paused,'' he said calmly.

He walked to the table, opened the \hl{Carnival Logic Binder}, and began reading aloud—not to shame, but to restore.

``The \hl{silver prize chest},'' he read, ``must always stay on the \hl{RIGHT} side of the main stage.''

He turned, lifted the chest with the help of two staff members, and returned it to its rightful place.

Next, he approached the bins.

``The \hl{red bin},'' he continued, ``holds only \hl{green tickets}. The \hl{blue bin} holds only \hl{yellow tickets}.''

Slowly and carefully, Vale returned the tickets to their proper bins. Then, in full view of the staff, he counted.

``One… two… three…''

When he finished, he announced clearly:

``\hl{Fourteen tickets total. Eight green. Six yellow}.''

Only then did he close the binder.
The correction complete, the rain passed, and the carnival resumed.

As evening fell, lanterns lit the grounds and the final ceremony began. Vale stood once more at center stage, his \hl{white tuxedo jacket} and \hl{black bowtie} unchanged. Iris joined him, still wearing her \hl{blue gloves}, smiling with quiet confidence. Pogo waved to the crowd, his \hl{red scarf} bright under the lights.

On the \hl{RIGHT} side of the stage, the \hl{silver prize chest} gleamed. The ribbon winners stood proudly—\hl{blue ribbon} for the kite contest, \hl{red ribbon} for the drum contest. Backstage, the bins remained untouched, holding exactly what they should.

Vale closed the ceremony with a nod.
The audience cheered, unaware of how close chaos had come—and how firmly logic had held.
The \hl{Carnival Logic Binder} was returned to its shelf, ready for the next year, its rules once again perfectly matched by reality.

\end{StoryCard}
\end{minipage}
\vspace{-0.6em}
\caption{A single-page story card used in our long-horizon constraint stress test. Highlighted phrases indicate invariant rules (character attributes, object placement, ticket inventory, and ribbon-to-contest mapping) and the explicit violation-and-correction episode.}
\label{fig:story_card_long}
\end{figure*}
\clearpage

\end{document}